\documentclass[preprint,11pt]{elsarticle}
\usepackage[utf8]{inputenc}
\usepackage[T1]{fontenc}
\usepackage{amsmath}

\usepackage{amsfonts}
\usepackage{amssymb}
\usepackage{graphicx}
\usepackage{vmargin}
\usepackage{float}
\usepackage{hyperref}
\setmarginsrb{2.5cm}{2.5cm}{2.5cm}{2.5cm}{0cm}{0cm}{0cm}{1cm}
\usepackage{xcolor} 
\usepackage{listings}
\usepackage{gensymb}
\usepackage{eurosym}
\usepackage{caption}
\usepackage{subcaption}
\usepackage{xfrac}
\usepackage{array}
\usepackage[linesnumbered,lined,boxed,commentsnumbered]{algorithm2e}
\usepackage{tikz}
\usepackage{changepage}
\usepackage{tablefootnote}

\begin{document}

\begin{frontmatter}
\title{A two-level machine learning framework for predictive maintenance: comparison of learning formulations}
\tnotetext[t1]{The research project was funded by the Walloon region under the BiDMed project.}
\author[add1]{Valentin Hamaide}
\author[add2]{Denis Joassin}
\author[add2]{Lauriane Castin}
\author[add1]{François Glineur}

\address[add1]{ICTEAM Institute, UCLouvain, Louvain-la-Neuve, Belgium}
\address[add2]{Ion Beam Applications, Louvain-la-Neuve, Belgium}

\begin{abstract}
\noindent Predicting incoming failures and scheduling maintenance based on sensors information in industrial machines is increasingly important to avoid downtime and machine failure. Different machine learning formulations can be used to solve the predictive maintenance problem. However, many of the approaches studied in the literature are not directly applicable to real-life scenarios. Indeed, many of those approaches usually either rely on labelled machine malfunctions in the case of classification and fault detection, or rely on finding a monotonic health indicator on which a prediction can be made in the case of regression and remaining useful life estimation, which is not always feasible. Moreover, the decision-making part of the problem is not always studied in conjunction with the prediction phase. This paper aims to design and compare different formulations for predictive maintenance in a two-level framework and design metrics that quantify both the failure detection performance as well as the timing of the maintenance decision. The first level is responsible for building a health indicator by aggregating features using a learning algorithm. The second level consists of a decision-making system that can trigger an alarm based on this health indicator. Three degrees of refinements are compared in the first level of the framework, from simple threshold-based univariate predictive technique to supervised learning methods based on the remaining time before failure. We choose to use the Support Vector Machine (SVM) and its variations as the common algorithm used in all the formulations. We apply and compare the different strategies on a real-world rotating machine case study and observe that while a simple model can already perform well, more sophisticated refinements enhance the predictions for well-chosen parameters.
\end{abstract}
\end{frontmatter}


\section{Introduction}
\noindent Predictive maintenance (PdM), or condition-based maintenance, consists of recommending maintenance decisions based on the information collected through condition monitoring, usually in the form of time series. It is usually formulated in one of the two following ways: i) detecting that the machine under monitoring has entered a faulty state, and therefore predicting that a failure is coming, or ii) predicting the remaining useful life (RUL) of the machine. In the scientific literature, those two approaches are referred to as \emph{diagnostics} and \emph{prognostics} respectively. \emph{Prognostics} is defined by Jardine et al. \cite{jardine2006review} in their review as "to predict faults or failures before they occur" and \emph{Diagnostics} as "focusing on detection, isolation and identification of faults when they occur" or as "a procedure of mapping the information obtained in the measurement/features space to machine faults in the fault space", i.e. pattern recognition. However, in this paper, those two concepts are not differentiated as our purpose is to schedule a maintenance procedure based on the current monitoring information, regardless of the type of method used. We propose a framework divided into two levels, the first level consisting of a machine learning model mapping a set of features into a health indicator and the second level being responsible for the actual decision making, where an alarm is raised if the health indicator of the first level crosses a threshold whose value can be optimized.\\
\\
Predictive maintenance methods can be categorized into model-based, statistical and machine learning approaches. In this paper, we will focus on machine learning (ML) approaches. For surveys on model-based and statistical methods, the reader can refer to \cite{gao2015survey, gao2015survey2}.
\subsection{ML based approaches for predictive maintenance}
\noindent Machine learning techniques applied to fault diagnosis were recently reviewed in \cite{lei2020applications}. Those techniques consist of designing machine learning models to establish the relationship between selected features and the health state of machines. In \cite{lei2020applications}, those methods are described chronologically, starting with traditional ML theories that include data collection, feature extraction and selection and health state recognition with models such as artificial neural networks (ANN), support vector machine (SVM) and k-nearest neighbors (KNN). In the present, deep learning methods such as convolutions neural network (CNN), auto-encoders (AE) or deep belief networks are used, which are capable of learning both the task and the feature space simultaneously. A common assumption of those methods is the availability of sufficient labelled data, which is usually hard to obtain in real-world engineering scenarios \cite{lei2020applications}. A more reasonable assumption would be a case where data has been collected for a few machines that have gone through a failure (or at least a deteriorate state) and underwent a corrective maintenance as well as for some other machines that have been replaced without failure (preventive maintenance). In that case, although different health states are available in the data, an exact labelling is not available since one does not necessarily know when a machine has entered a faulty state. In this paper, we investigate different formulations on how to express such problems with a machine learning formulation with variations of the support vector machine (SVM) algorithm. The reason for choosing this algorithm over deep learning approaches is threefold. First, we want to isolate the formulation or labelling scenario as much as possible from the algorithm for our comparison. Second, our case study involves only a few instances of failures and SVM tends to have good generalization capabilities even with few instances \cite{cervantes2020comprehensive}. Finally, the goal of this paper is not to seek for the best ML algorithm but to compare learning formulations.

\subsection{Classification}
\noindent In the case of \emph{classification}, we seek to find whether or not a machine will go to a failure state within a given time-window. The duration of the time-window is then a parameter that has to be chosen by the user. Surprisingly, very few papers treat the case of finding incoming faults in machines with a labelling purely based on failure time (e.g. the last 5 days are labelled as faulty). One of the papers that is using this kind of approach is \cite{susto2014machine} where the authors use a binary classification algorithm with different horizons to define the faulty class and select the one which optimize a custom cost that is a trade-off between fault detection and unexploited life (i.e. replacing the component too soon and not exploiting the full life of the component).

\subsection{Anomaly detection}
\noindent Besides supervised learning, \emph{anomaly detection} can be used for fault diagnosis. In the simple univariate case, a threshold can be learned on a specific feature which is considered anomalous when the threshold is reached (e.g. in \cite{wang2002model}). In the multivariate case, a Gaussian distribution can be fitted on healthy data and the Mahalanobis distance used as an indicator of health. This is the approach taken by \cite{jin2016anomaly, wang2016two}. Another option is to use the one-class SVM algorithm with a model learned on healthy data such as the authors did in \cite{fernandez2013automatic, mahadevan2009fault}. Deep learning methods based on autoencoders were also applied for anomaly detection in the context of fault diagnosis in \cite{reddy2016anomaly}.

\subsection{Regression and remaining useful life estimation}
\noindent Prognostics, the second type of PdM approach, is usually implemented as a Remaining Useful Lifetime (RUL) estimation \cite{jardine2006review}, that is predicting at each time step the number of hours/days/cycles remaining before the machine goes into failure. Various review papers have been published over the past 15 years on the subject \cite{jardine2006review, heng2009rotating, sikorska2011prognostic, lee2014prognostics, si2011remaining, lei2018machinery} and most of the papers do not differentiate prognostics from RUL prediction.\\
\\
RUL estimation is a regression problem. However, it is traditionally not implemented as a regular supervised learning problem where a mapping is learned between the sensor inputs and the RUL. The standard approach is to extrapolate the trend of one or several health indicators previously extracted from the data via signal processing or machine learning methods until it reaches a predefined threshold. A systematic approach to RUL prediction in that sense is presented in \cite{lei2018machinery}. Examples of machine learning approaches using this kind of framework can be found in \cite{soualhi2014bearing}. However, a few papers try to directly map the inputs to the actual RUL. This is the case in \cite{khelif2016direct} where the authors apply a support vector regression algorithm to the NASA's Turbofan engine degradation dataset (CMAPSS) \cite{saxena2008damage} and more recently with deep learning such as in \cite{li2018remaining, wu2018remaining} on the same dataset. However, this dataset is a simulated dataset with manually introduced failures \cite{lei2018machinery}, which can be quite different from real-life scenarios. In practice, it is usually hard to estimate the RUL in engineering scenarios, especially in the absence of clear degradation trends. In this paper, a mapping is learned between the feature space and the RUL; however, the final goal is not the RUL itself but the decision taken upon it, i.e. when an alarm for replacement should be raised based on the predicted RUL. Other regression approaches using a different labelling scheme are also tested in the paper and discussed in Section \ref{sec:regression}.

\subsection{Decision making}
\noindent The use of the ML algorithm in the first level of our framework has a different purpose from what is commonly done within the predictive maintenance community. Indeed, we are interested to extract a single continuous health indicator from the learning algorithm rather than identifying a certain state in the case of classification, an outlier in the case of anomaly detection or a RUL in the case of regression. This first level is completely unaware of time, its main purpose is pattern recognition. The resulting health indicator is then fed to the second level of the framework which is responsible for taking a decision for maintenance based on a smart aggregation of the past values of the health indicator.\\
\\
The rest of the paper is structured as follows: in Section \ref{sec:framework}, the framework for predictive maintenance is outlined. In Section \ref{sec:formulations}, various formulations for the learning problem are described. The decision-making process is presented in Section \ref{sec:decision}. In Section \ref{sec:validation}, the method of validation of the models is presented. The different methods are tested on a rotating machine case study in Section \ref{sec:application} and finally Section \ref{sec:conclusion} concludes.

\section{Predictive maintenance framework}
\label{sec:framework}
\begin{figure}[h]
    \centering
    \includegraphics[scale=0.59]{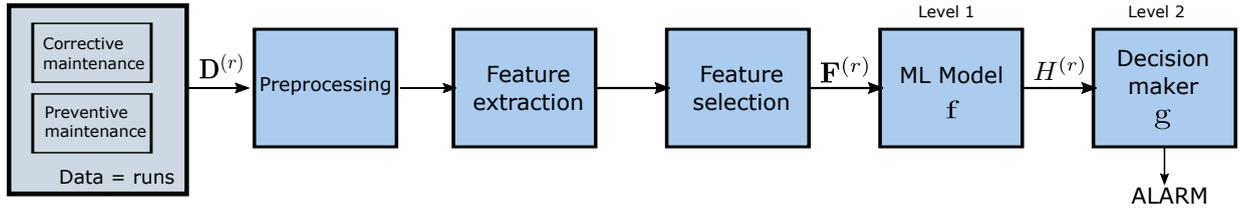}
    \caption{Predictive maintenance framework}
    \label{fig:framework}
\end{figure}
\begin{figure}[t]
    \centering
    \includegraphics[scale=0.55]{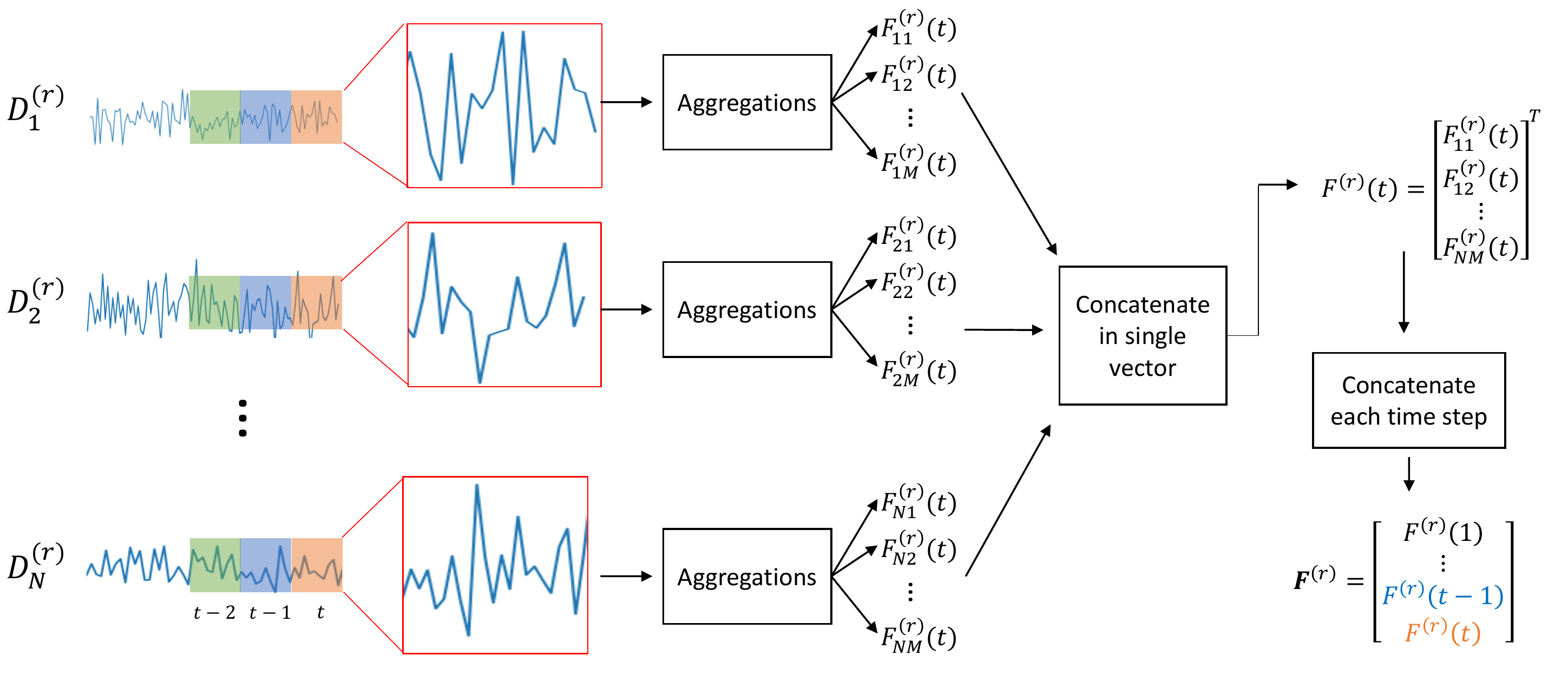}
    \caption{Feature design}
    \label{fig:features}
\end{figure}

\noindent A flowchart for our predictive maintenance framework is depicted in Figure \ref{fig:framework}. First, data need to be acquired via sensor measurements in the form of time-series. The purpose of the data acquisition step is to obtain data from several machines that got a corrective maintenance, i.e. that went through failure (or at least until a deteriorated state) but also possibly from machines that were preventively replaced and did not go into a failure state. We call the data collected from a particular machine from the installation to the replacement (with or without failure) a \emph{run}. Let us define $\mathbf{F}^{(r)} \in \mathbb{R}^{n_r\times N}$, the feature matrix of the run $r$, where $n_r$ is the number of samples in the time series and $N$ the number of features. The dataset consists of $R$ runs, i.e. $r=1,...,R$. The notation $\mathbf{F}^{(r)}(t)$ is used to access the $t^{th}$ sample in the matrix. We also define $x = \left[ F^{(r)} \right]_{r=1,...,R}$ the merged feature matrix of size $\sum_r n_r \times N$, and $x_i \in \mathbb{R}^N$ refers to sample number $i$ (regardless of time) of the dataset. Some preprocessing is then achieved on the data collected, which can include filling missing values, removing sensor faults, normalization, etc.\\
\\
The next step consists of extracting meaningful information from the raw sensor measurements $\mathbf{D}^{(r)}$. This is called feature extraction by the machine learning community while sometimes referred as health indicator construction \cite{lei2018machinery} in the predictive maintenance literature. In the context of time-series data, it consists of designing features aggregating sensor inputs that summarize a certain time period rather than a point in time. Those aggregations can be done in the time-domain, frequency-domain or time-frequency domain. A description of the feature design is shown in Figure \ref{fig:features}. In the case of vibration data, the reader can refer to \cite{wang2017prognostics} for a review of the common features that are used for predictive maintenance.\\

\begin{figure}[t]
\centering
\begin{subfigure}{.55\textwidth}
  \centering
  \includegraphics[width=1\linewidth]{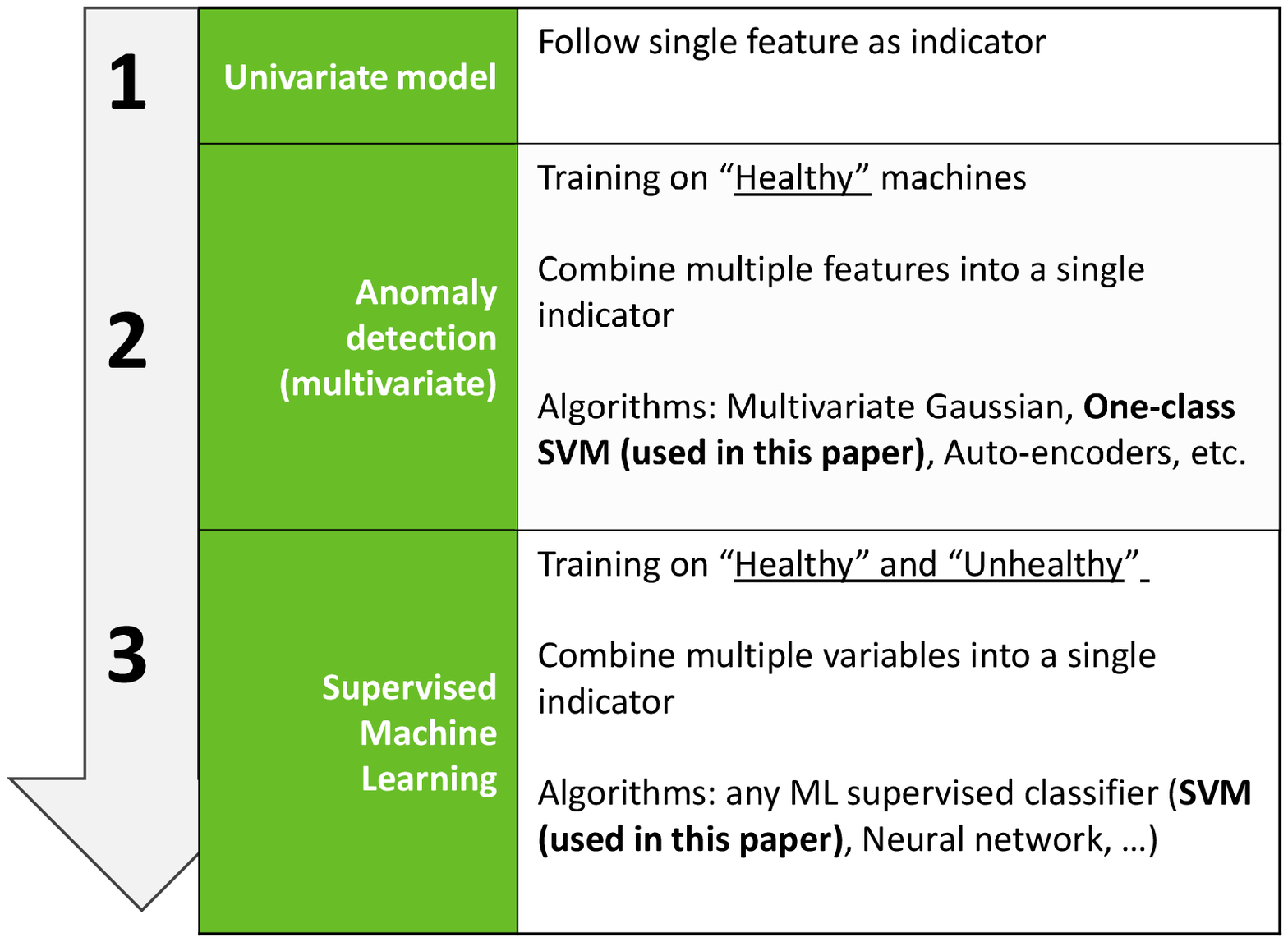}
  \caption{First level: Three refinement of the ML model}
  \label{fig:models}
\end{subfigure}%
\begin{subfigure}{.45\textwidth}
  \centering
  \includegraphics[width=1\linewidth]{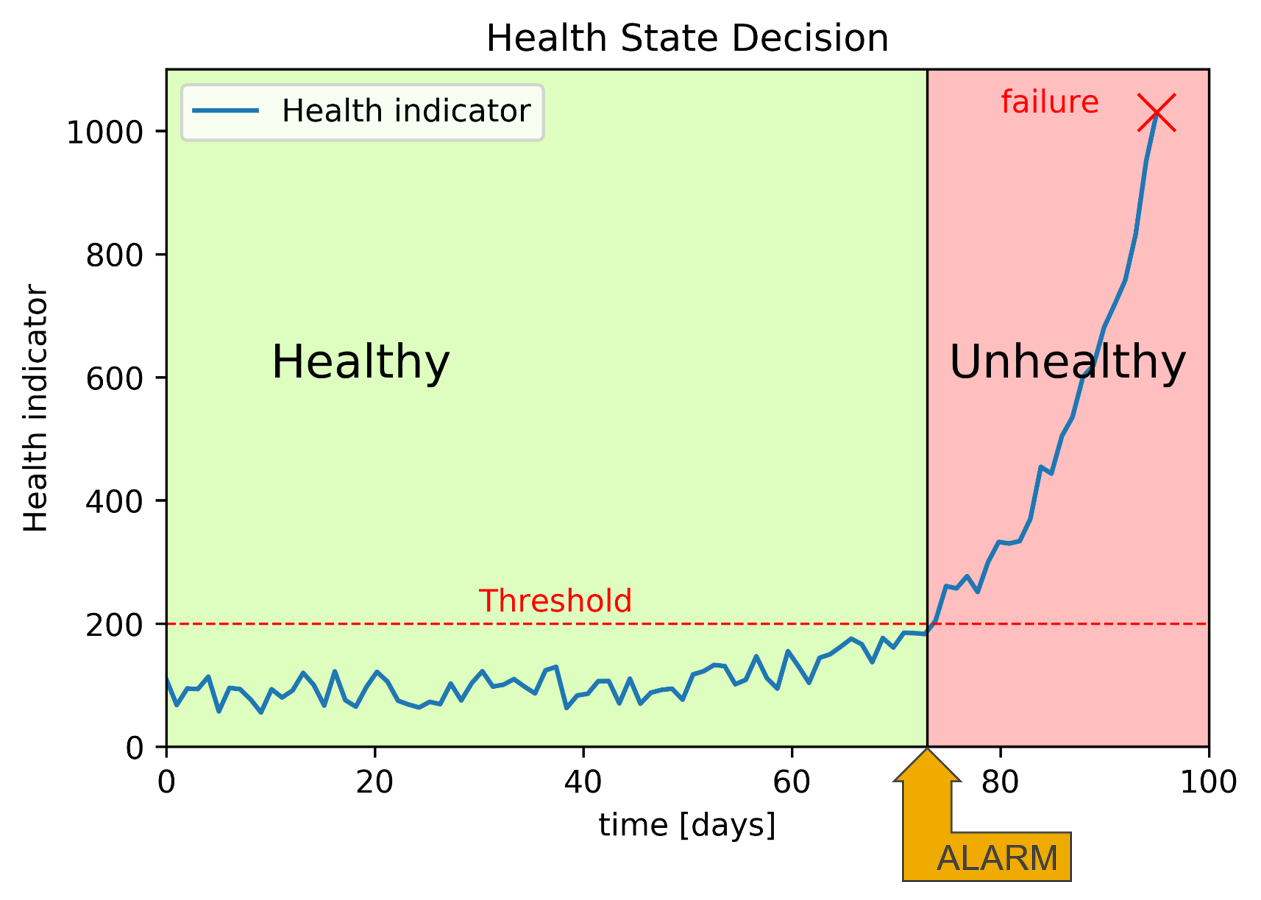}
  \caption{Second level: Decision maker}
  \label{fig:decision}
\end{subfigure}
\caption{Two-level predictive maintenance: (a) a model translates a set of features into a single indicator (b) a decision is made to trigger an alarm when the health indicator exceeds an optimized threshold (in this case $g$ is the identity function).}
\label{fig:predictive_maintenance}
\end{figure}

\noindent A large number of features can be initially produced by the feature extraction step. Selecting features targets two goals: removing uninformative features and reducing the size of the problems to mitigate some possible overfitting and improve computational efficiency. Two main strategies can be used to select features: a wrapper approach and a filter approach. The wrapper approach selects the features according to the model in an iterative loop where features are added or removed according to the score obtained with the model. The filter approach is model agnostic and selects the features \emph{a priori} according to a certain criterion. Since we are comparing different models, we will choose a filter approach so that we have a set of features common in all experiments. However, the criteria to select features for the filter approach should not be supervised, since different labelling strategies will be tested and choosing one would unfairly favor the algorithm that uses this particular labelling. Instead, we use an unsupervised minimum redundancy maximum relevance feature selection for predictive maintenance which uses a combination of three prognostic metrics to quantify the relevance of a feature while minimizing redundancy between features \cite{hamaide2021unsupervised}. The prognostic metrics evaluate what a good prognostic parameter should be, i.e. monotonic with respect to time, behaving in the same ways for different machines and easily separable between the starting and failure values. For details on the feature selection method, the reader can refer to \cite{hamaide2021unsupervised}.\\
\\
From the selected set of features, we wish to establish a causal relationship with the health state of the machine. If the current health state is considered faulty, an alarm should be triggered for repair or replacement. This is done in two steps. In a first step, the input features are fed to a model learned on previous machines, which outputs a single number at each time step (or a single time-series across time), i.e. $h^{(r)}(t)=f(\mathbf{F}^{(r)}(t))$. Then, in a second step, a decision is taken upon this output. An alarm is triggered if $g(h^{(r)}(t))$ exceeds (or goes below) a threshold whose value was optimized beforehand on different machines, where $g$ is a certain time-window transformation applied to the health indicator computed from the model. The function $g$ can be the identity function or a certain aggregation of the output across time such as a moving average or exponential moving average. This two-level predictive maintenance approach is depicted in Figure \ref{fig:predictive_maintenance}.\\


%

\noindent The next section compares the different strategies that map the set of features into a single health indicator, i.e. the ML model part in the diagram of Figure \ref{fig:framework}.

\section{First level: Problem formulations}
\label{sec:formulations}
\noindent In this section, we compare three different strategies to map a set of features into a health indicator. A first possibility is to simply follow a single feature, a second requires to train the algorithm on healthy data and consider as anomalous what deviates from the norm, and finally the most refined technique is to train on both healthy and unhealthy data with supervised learning algorithms, which can be formulated either as binary classification, multi-class classification or regression. As mentioned earlier, the machine learning models used is this paper are variation of the support vector machine algorithm (SVM). For the anomaly detection, the one-class SVM is used, for the classification and multi-class classification, the standard SVM is used and for the regression, the support vector regression (SVR) algorithm is used.

\subsection{Univariate model}
\noindent In the univariate case, a single feature is followed across time. The selected feature can be chosen according to engineering expertise or according to the maximal relevance score obtained by a feature selection algorithm. In our case, the relevance criterion is the average of three prognostic metrics: monotonicity, trendability, prognosability \cite{hamaide2021unsupervised}. Additionally, some aggregations can be performed on a certain time-window up to the current time instant, such as taking the moving average across several hours or even days if needed. More complex aggregations can also be performed such as exponential moving average, autoregressive models, etc. The health indictor computed from the ML model is then $$h^{(r)} = f_U\left(\mathbf{F}^{(r)}\right)$$ where $f_U$ is the model (any type of aggregation or simply the identity mapping) where $\mathbf{F}^{(r)}$ contains only one feature in this case.

\subsection{Multivariate anomaly detection}
\noindent When performing anomaly detection, the algorithm is only trained on healthy data. When testing on new data, a sample is either marked as inlier or outlier (i.e. anomalous). In practice, the decision is not binary but is taken based on a threshold for a decision function. For instance, in the case of anomaly detection based on a multivariate gaussian distribution fitted on healthy data, a decision to flag a sample as anomalous is taken if it is far from the fitted n-dimensional ellipsoid center. Usually, the distance measure chosen is the Mahalonibis distance between the sample and the ellipsoid center. The evolution of that distance across time will therefore be the health indicator on which a decision to trigger an alarm will be made. In our approach, the algorithm chosen for anomaly detection is the one-class SVM, rather than an approach based on the multivariate gaussian algorithm. Even though the Mahalanobis distance does not apply to the one class SVM, there is a similar idea of distance from a set of \emph{normal} samples.\\
\\
One-class SVM was proposed as an extension of the support vector machine to the case of unlabeled data \cite{scholkopf2001estimating}. It tries to estimate the distribution of the input data (considered healthy) by a simpler subset of the input space and estimates a function $f$ that is positive in that subset and negative on the complement. The corresponding model is formulated as follows:
\begin{eqnarray}
\label{eq:one_class}
\min_{w, b, \zeta} \frac{1}{2} \lVert w \rVert ^2 + \frac{1}{\nu \ell} \sum_{i=1}^{n} \zeta_i - b\\
\textrm{s.t. } (w^T \phi (x_i)) \geq b - \zeta_i, \\
\zeta_i \geq 0, i=1,.., n 
\end{eqnarray}
where $x_i$ are the training vectors, $w$ and $b$ the weights and bias for which we solve, $\zeta_i$ are the slack variables allowing some samples to be on the \emph{wrong} side of the hypersurface, $\phi(\cdot)$ is a non-linear mapping to allow for a non-linear boundary, $\ell$ is the number of samples in the training set and $\nu \in [0,1)$ is a hyperparameter representing an upper bound on the fraction of training errors and a lower bound of the fraction of support vectors. For our experiments, as well as the other SVM-based models, we used the \textit{sci-kit learn} implementation which is wrapper around the LIBSVM library \cite{chang2011libsvm}. The decision-making process is based on the distance to the hypersurface function, and the decision function is defined as
\begin{equation}
\label{eq:decision_svm}
    f_{\mathrm{1SVM}}(x)=\sum_{i=1}^n \alpha_i K(x_i, x) + \rho
\end{equation}
where $K(x_i,x) = \phi (x_i)^T\phi (x)$ is the kernel function and $\alpha_i$ and $\rho$ the dual variable and independent term of the optimization problem \eqref{eq:one_class}. The health indicator computed from the model is then $$h^{(r)} = f_{\mathrm{1SVM}}\left(\mathbf{F}^{(r)}\right)$$
The same notation will be used throughout the paper for all SVM-based models for variables, weights, slacks and kernels.

\subsection{Supervised learning}
\noindent Supervised learning refers to algorithms that learn a function mapping from a set of input variables, the features, to a corresponding output, the labels. This requires having the corresponding desired label for each input sample. Those labels can either be numeric values in case of regression or categorical variables in case of classification. In the context of predictive maintenance, those categorical variables can be the fact that a machine is in a healthy or unhealthy state or a certain type of faults on the machine. If the labels are known, for instance with a healthy machine and a faulty machine on a test bench, then the problem becomes simple and supervised learning is the way to go. However, labels are usually not available in real-case scenarios. Indeed, we do not necessarily know when exactly a machine enters a faulty state, even if the machine goes into failure at the end of the run. Instead, we use a labelling based on the remaining time before failure. In the case of classification and multi-class classifications, the labels are chosen somewhat arbitrarily by splitting the run into two or more time-periods representing healthy or unhealthy states. The chosen duration used to split the run into classes is a parameter that needs to be chosen by the user based on engineering expertise or tuned as an hyperparameter of the problem. In the case of regression, the time before failure can be directly used as the labelling but other possibilities exist and are described in Section \ref{sec:regression}.

\subsubsection{Binary classification}
\noindent In the classification approach, a run is divided into two classes: faulty (F) and non-faulty (NF) as follows:
$$y^{(r)}(t) =
    \begin{cases}
      NF & \text{if $t< T^{(r)} - w$}\\
      F & \text{if $t \geq T^{(r)} - w$}
    \end{cases}   $$
where $T^{(r)}$ is the duration of the run $r$ and $w$ is a parameter to choose for the faulty state duration. We thus have to choose a duration \emph{a priori} taking into account that a $w$ too small will lead to a late detection while a $w$ too big could lead to too early detection (and therefore unexploited lifetime). In the case of a machine preventively replaced (no actual failure at the end of the run), the entire run is marked as $NF$.\\
\\
The classification algorithm used here is the well-known support vector machine algorithm \cite{boser1992training} which finds the hypersurface that separates the classes with maximal margin by solving the following optimization problem:
\begin{eqnarray}
\label{SVM_opti}
\min_{w, b, \zeta} \frac{1}{2} \lVert w \rVert ^2 + C \sum_{i=1}^{n} \zeta_i\\
\textrm{s.t. } y_i (w^T \phi (x_i) + b) \geq 1 - \zeta_i, \\
\zeta_i \geq 0, i=1,.., n \label{SVM_opti3}\end{eqnarray}
where $y_i \in \lbrace -1, 1\rbrace^n$ are the target values (where $F$ is mapped to 1 and $NF$ to -1) and $C$ is an hyperparameter representing the trade-off between the margin width and the sum of the slack variables ($\sum_i \zeta_i$).\\
\\
The decision function is based on the distance to the hypersurface, that is defined as
\begin{equation}
\label{eq:decision_svm}
    f_{\mathrm{SVM}}(x)=\sum_{i=1}^n y_i \alpha_i K(x_i, x) + \rho
\end{equation}
and the computed health indicator is thus $$h^{(r)} = f_{\mathrm{SVM}}\left(\mathbf{F}^{(r)}\right)$$

\subsubsection{Multi-class classification}
\noindent In the multi-class classification approach, a run is separated into more than two classes. Each class represents a different non-overlapping time window between the beginning of the run and the failure. Mathematically, the labels are defined as $$y^{(r)}(t) =
    \begin{cases}
      NF & \text{if $t < T^{(r)} - w_1$}\\
      F_1 & \text{if $T^{(r)} - w_{1} \leq t < T^{(r)} - w_2$} \\
      F_2 & \text{if $T^{(r)} - w_{2} \leq t < T^{(r)} - w_3$} \\
      \vdots \\
      F_{N-1} & \text{if $t \geq T^{(r)} - w_{N}$}
    \end{cases}   $$
where $NF$ is considered the healthy class and $F_i, i=1,...,N-1$ are considered the $N-1$ faulty classes with $w_1>w_2>...>w_N$ and increased level of fault severity. In the case of a machine preventively replaced (no failure at the end), the entire run is marked as $NF$.\\
\\
The multi-class classification approach is similar to the binary approach. The SVM algorithm is still used but instead of solving one optimization problem as in (\ref{SVM_opti}-\ref{SVM_opti3}), we solve $N$ optimization problems where $N$ is the number of classes. We use a one-versus-rest strategy where we train a single classifier per class, with the samples of that class being positive ($+1$) and all the rest being negative ($-1$) and thus keep the same formulation as in (\ref{SVM_opti}).\\
\\
Since there are $N$ optimization problems, there are $N$ decision functions such as in (\ref{eq:decision_svm}). The decision functions are defined as $$h_j^{(r)} = f_{\mathrm{SVM}}^j\left(\mathbf{F}^{(r)}\right) \quad \text{for } j=1,...,N$$


\subsubsection{Regression}
\label{sec:regression}
\noindent In the regression approach, three labelling scenarios are tested. The first approach is to directly use the remaining useful life (RUL) as labels. In the second approach, instead of using absolute times, relative times are used with the percentage of life used as labels. Finally, a third approach tries to mimic the intuition that a machine is stable in the beginning of its life and deteriorates more and more starting some time before the failure with a piecewise linear function. We refer to this approach as \emph{ReLu}, an analogy to the rectified linear unit in machine learning due to the shape of the labelling function. The three labelling approaches are detailed in Table \ref{tab:regression}. Those labelling strategies are only valid for corrective maintenance. Indeed, runs of machines preventively replaced have to be disregarded in case of RUL or percentage of life strategies. For the ReLu strategy, runs of machines preventively replaced can be kept and labelled as zero across the entire life span of those machines.\\

\begin{table}[h]
    \centering
    \begin{tabular}{|m{0.1\textwidth}|m{0.262\textwidth}|m{0.262\textwidth}|m{0.262\textwidth}|}
    \hline
         & \textbf{RUL} & \textbf{Percentage of life} & \textbf{ReLu} \\\hline
        Labelling & \begin{equation*}
\label{eq:RUL_label}
    h^{(r)}(t) = \frac{T^{(r)} - t}{D}
\end{equation*} with $D$ a normalizing constant & \begin{equation*}
    h^{(r)}(t) = \frac{t - t_0^{(r)}}{T^{(r)} - t_0^{(r)}}
\end{equation*}
where $t_0$ is the start time and $T$ the end time. & \begin{eqnarray*}
    h^{(r)}(t) =\\ \begin{cases}
      0 & \text{if $t<T - t_d$}\\
      \frac{T-t}{t_d} & \text{if $t \geq T - t_d$}
    \end{cases}
\end{eqnarray*}
where $T$ the end time and $t_d$ is a supposed start of deterioration fixed for training.\\\hline
        Function shape & \begin{tikzpicture}

\draw (0,1) -- (2,0);
\node at (-0.2,1) {\footnotesize T};
\node at (2.2,0) {\footnotesize 0};

\end{tikzpicture} & \begin{tikzpicture}

\draw (0,0) -- (2,1);
\node at (-0.2,0) {\footnotesize 0};
\node at (2.2,1) {\footnotesize 1};

\end{tikzpicture} & \begin{tikzpicture}

\draw (0,0) -- (1.5,0) -- (2,1);
\node at (-0.2,0) {\footnotesize 0};
\node at (2.2,1) {\footnotesize 1};

\end{tikzpicture}\\\hline
        Decision & $g\left(h^{(r)}\right)<L$ & $g\left(h^{(r)}\right)>L$& $g\left(h^{(r)}\right)>L$ \\\hline
    \end{tabular}
    \caption{Labelling strategies for regression \& decision making}
    \label{tab:regression}
\end{table}

\noindent Support vector regression (SVR) is the common regression algorithm used for all labelling scenarios. It is a variation of the SVM performing a regression based on the concept of support vectors \cite{drucker1997support}. The idea is to find a function $f(x)$ that has at most an $\varepsilon$ deviation from the targets $y_i$ by solving the following optimization problem
\begin{eqnarray}
\label{eq:SVR_opti}
\min_{w, b, \zeta, \zeta^*} \frac{1}{2} \lVert w \rVert ^2 + C \sum_{i=1}^{n} (\zeta_i + \zeta_i^*) \\\textrm{s.t. } y_i - w^T \phi (x_i) - b \leq \varepsilon + \zeta_i,\\
w^T \phi (x_i) + b - y_i \leq \varepsilon + \zeta_i^*,\\
\zeta_i, \zeta_i^* \geq 0, i=1,..,n\
\end{eqnarray}
where $y_i$ are the targets values, $\zeta_i$ and $\zeta_i^*$ are the slack variables allowing some samples to be outside of the tube of radius $\varepsilon$ centered around the function and $C$ is an hyperparameter representing the trade-off between the flatness of $f$ and the sum of deviations larger than $\varepsilon$ (that is $\sum_i \zeta_i$+$\zeta_i^*$). For more information on the SVR and how this optimization problem can be solved efficiently, the reader can refer to \cite{smola2004tutorial}.\\
\\
The decision function of the SVR is defined as
\begin{equation}
\label{eq:decision_svr}
    f_{\mathrm{SVR}}(x) = \sum_{i=1}^n (\alpha_i - \alpha_i^*) K(x_i, x) + \rho
\end{equation}
where $\alpha_i$, $\alpha_i^*$ and $\rho$ are the dual variables and independent term of the optimization problem (\ref{eq:SVR_opti}). The computed health indicator (HI) is thus $$h^{(r)} = f_{\mathrm{SVR}}\left(\mathbf{F}^{(r)}\right)$$


\section{Second level: Decision making}
\label{sec:decision}
\noindent The purpose of the second level of the predictive maintenance framework is the make a decision on the health indicator computed from the ML model. Let $h^{(r)}=f(\mathbf{F}^{(r)})$ be the HI computed from the model, the decision maker will trigger an alarm for replacement if $z=g(h^{(r)})>L$ where $g$ is a function of time-series $h^{(r)}$ and $L$ is a predefined threshold. In the simplest form, $g$ is the identity function, or the negative function $g(h^{(r)})=-h^{(r)}$ in case $L$ is a lower bound. When the current output is above (or below) the threshold $L$, an alarm is raised. The function $g$ can also be an aggregation of past values such as a moving average or an exponential moving average. In case $g$ is a moving average, it is defined as
\begin{equation}
    g(y) = \frac{h(t)+h(t-1)+...+h(t-H)}{H}
\end{equation}
where $H$ is the horizon selected for the aggregation. In case $g$ is an exponential moving average, the aggregation is defined recursively as
\begin{eqnarray}
z_0 &=& h_0 \\
z_t &=& \alpha h_t + (1-\alpha)h_{t-1}
\end{eqnarray}
The difference with the simple moving average is that in this case, the window-size is infinite but the weights are exponentially decreasing. However, the $\alpha$ parameter can be tuned to be interpreted approximately as a H-hour moving average when computed as 
\begin{equation}
\label{eq:alpha}
    \alpha=\frac{2}{H+1}
\end{equation}
where $H$ is the horizon\footnote{This is an approximation of of the formula $\alpha=1-\exp(\log(1-p)/H)$ where $p=0.86$ is the contribution of the window on the moving average}. For instance, if $H=12$ and the time between two consecutive sample is 1 hour, $\alpha=\frac{2}{12+1}\approx 0.1538$ and is interpreted as a 12-hour moving average.\\
\\
In the case of binary classification, an alarm would be triggered if $z(t)>L$ where $L$ is a previously optimized threshold. For regression, the idea is similar and the decision functions are detailed in Table \ref{tab:regression}. For the multi-class classification, since there are multiple decision functions, the process has to be slightly adapted. An alarm is triggered if the value of the decision function of one of the faulty classes ($F_1,...,F_{N-1}$) is higher than the healthy class $NF$. Mathematically, we trigger an alarm at sample $i$ if any $j\neq 1$ satisfies
\begin{equation}
    g\left(f_{\mathrm{SVM}}^{F_j}\left(\mathbf{F}^{(r)}\right)\right) > g\left(f_{\mathrm{SVM}}^{NF}\left(\mathbf{F}^{(r)}\right)\right)
\end{equation}
where $F_j$ is the decision function associated to class $F_j$, $j=1,...,N-1$, $NF$ the decision function associated to the healthy class and $g$ is the function applied to the health indicator.

\section{Assessing the predictive performance of models}
\label{sec:validation}
\subsection{Scoring}
\noindent The end goal of a predictive maintenance application is to help on the decision to replace or repair the machine under monitoring at the right time. The right time may vary between applications but is usually a trade-off between detecting failure and limiting false alarms. As a first goal, we want to check whether or not a fault (in the case of a run with failure) can be detected and minimize false alarm. As a second goal, we want to score the timing of the alarm. This concept can be translated into two metrics. We define a false positive as an alarm that was triggered too early, in our case more than 15 days in advance. A true positive is defined as an alarm raised between 0 and 15 days in advance. While true positives are only relevant for corrective maintenance, false positives are relevant for both corrective and preventive maintenance. Indeed, an alarm raised at any time for a machine which was preventively replaced is considered a false positive. We thus define two metrics, the false positive rate:
\begin{equation}
    \text{FPR} = \frac{\text{FP}}{C + P}
\end{equation}
and the true positive rate
\begin{equation}
    \text{TPR} = \frac{\text{TP}}{C}
\end{equation}
where $C$ and $P$ are the numbers of corrective and preventive runs respectively, and FP and TP the numbers of false positives and true positives respectively. Note that we do not assess each individual prediction for each time step, but only evaluate the quality of the first trigger for each run (hence the earliest for that run). \\
\\
We combine those two metrics into a single metric by taking their harmonic average, similarly that we would do to compute the $F_1$ score between precision and recall in a conventional classification scenario. Moreover, we add a $\beta$ parameter, controlling the importance of the $\text{TPR}$ over the $1-\text{FPR}$. We call this metric, the $F_{\text{score}}$ and define it as
\begin{equation}
\label{eq:f_score}
    F_{\text{score}} = (1 + \beta^2) \cdot \frac{(1-\text{FPR)} \cdot \mathrm{TPR }}{\beta^2 \cdot (1-\mathrm{FPR}) + \mathrm{TPR}}
\end{equation}
Since it is also important to score the quality of timing at which the alarm is raised, another metric called the business metric ($B_{\text{score}}$) is defined. It provides a score between zero and one (higher is better) computed from a piecewise-linear function of the number of days between the prediction and the actual failure: \vspace*{.3cm}

\begin{minipage}{0.4\linewidth}
\label{fig:business_metric}
\hspace{-0.6cm}
\includegraphics[width=1.15\linewidth]{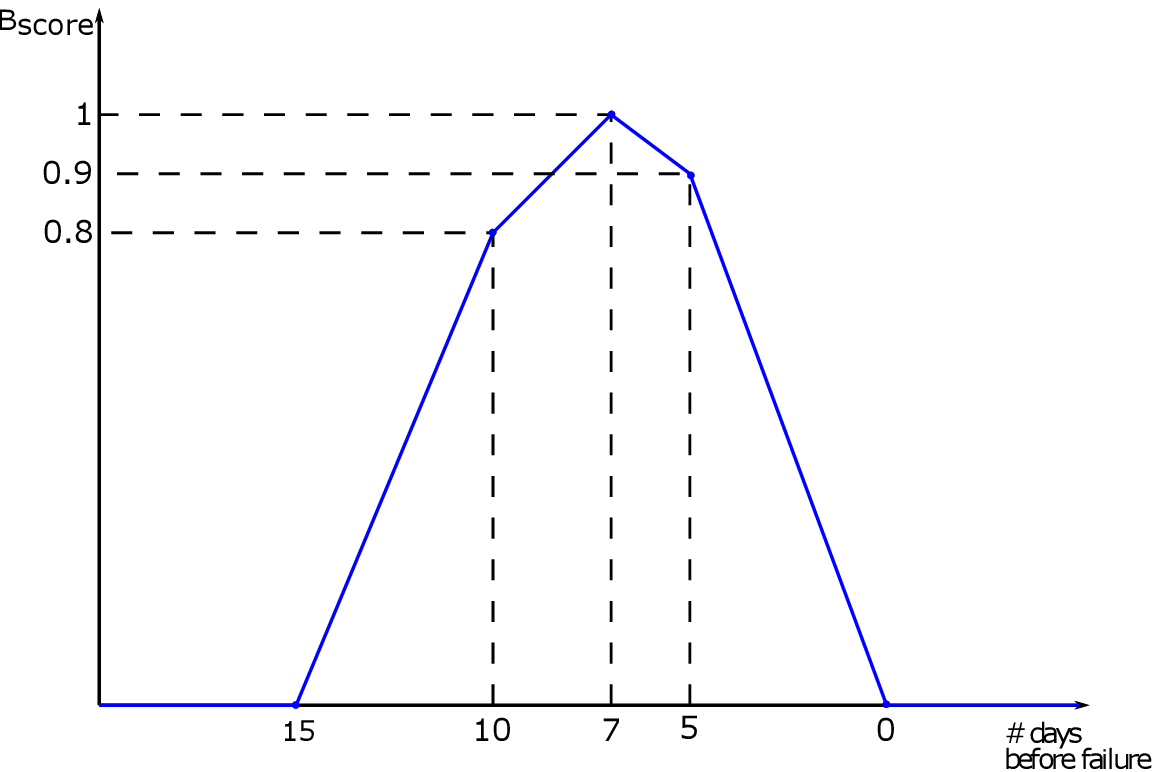}
\end{minipage}
\begin{minipage}{0.55\linewidth}
\small
\vspace{0.3cm}
\begin{itemize}
    \item A prediction 7 days ahead is considered optimal and gives a perfect score.
    \item Predictions between 7 and 0 days assign a score that decreases linearly to zero (with a slightly lower slope between 7 and 5 days).
    \item Predictions ranging from 7 to 15 days are assigned a score linearly decreasing from 1 to 0 (with a slightly lower slope between 7 and 10 days).
    \item Predicting a failure more than 15 days in advance leads to a zero score to reflect the unexploited lifetime.
\end{itemize}
\end{minipage}%
\vspace{0.5cm}

\noindent Note that this business score is only applicable for corrective maintenance. While $F_{\text{score}}$ is a single score resulting from an ensemble of runs, $B_{\text{score}}$ is defined per run and one must take the average across all corrective maintenance to obtain a single score, i.e. $\overline{B_{\text{score}}} = \sum_r \frac{B_{\text{score}}^{(r)}}{C}$.
Finally, we combine the $F_{\text{score}}$ and the business score into a single score that takes into account both the corrective and preventive maintenance runs. We define $\alpha\in[0,1]$ as the weight associated to the $F_{\text{score}}$ and $1-\alpha$ the weight associated to $\overline{B_{\text{score}}}$. Since the $F_{\text{score}}$ is used for both corrective and preventive maintenance while $B_{\text{score}}$ is only applied to corrective maintenance, we must further multiply $1-\alpha$ (the weight associated to $B_{\text{score}}$) by the ratio of corrective maintenance over the the number of runs, i.e. $\frac{C}{C+P}$. In the end, the final score is defined as

\begin{equation}
\label{eq:final_score}
    \frac{\alpha}{(1-\alpha)\frac{C}{C+P} + \alpha} F_{\text{score}} + \frac{(1-\alpha)\frac{C}{C+P}}{(1-\alpha)\frac{C}{C+P}+\alpha}\overline{B_{\text{score}}}
\end{equation}

\subsection{Cross validation}
\begin{figure}[h]
    \hspace{-.5cm} \includegraphics[scale=0.42]{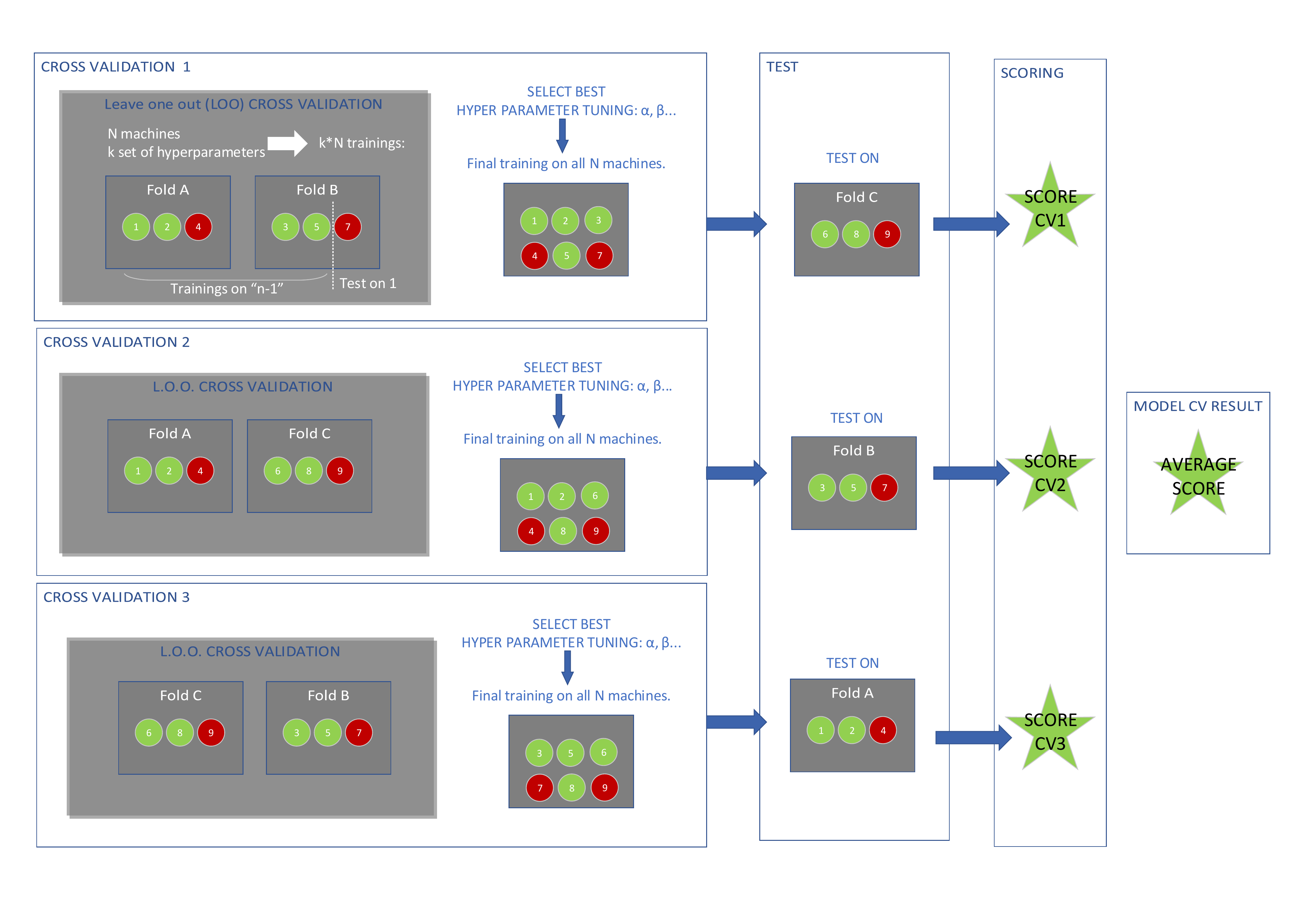}
    \vspace{-1cm}
    \caption{Example of double cross validation process for 9 machines subdivided into 3 folds.}
    \label{fig:cv}
\end{figure}

\noindent In order to obtain an unbiased estimation of the performance of the algorithms as well as tuning the different hyperparameters of the machine learning model and the threshold for the decision making, one must cross-validate the results. The usual way to validate a model in the context of machine learning is to split the data into a training, validation and test set. The training set is used to train the algorithm, the validation set is used to tune the hyperparameters of the model and the test set, a completely independent set, is used to assess the final prediction on unseen data. \\
\\
In the context of predictive maintenance, some precautions are necessary. Splitting the data into training, validation and test sets cannot be done in a completely random way. Recall that prediction occurs in a continuous fashion along the time series, i.e. we classify or predict at each time step. Hence, due to the temporal nature of the prediction task, data are correlated in time and one cannot use information learned in the future to predict the past or the present. Therefore, the training set should not contain data that are further in time than the validation and test set. An even better practice is to split data per run, meaning that data from a particular machine cannot be split among different sets. This is the approach that we take. Some machines are assigned to the training set, some to the validation set and the rest to the test set. \\
\\
However, in most real-life applications, machine runs-to-failure are scarce. Hence, it is difficult to build a sufficiently large (in terms of number of corrective maintenance) training set or test set. Therefore, we perform a cross-validation, where a fold is defined as a run. However, a single layer of cross-validation, for example a leave-one-out cross-validation, is still biased because the hyperparameter optimization has \emph{seen} all data. Since data are scarce, it is not an option to leave a few runs as test set as they would most likely not represent very well the behaviors of all machines. Instead, we perform what we call a double cross validation. The framework for double cross-validation is outlined in Figure \ref{fig:cv}.
This double cross-validation consists of two levels of cross-validations. In the inner loop a leave-one-out cross validation is performed to tune the hyperparameters of the ML model as well as tune the threshold for the decision-making. Then, a model is trained on all data of the inner loop (training set + validation set) with the best set of hyperparameter found in that inner loop. \\
\\
In the outer loop, the model is tested on an independent set of runs. Then, a new test set is selected at the next outer loop. In the end, we obtain not one but multiple models (with possibly different set of hyperparameters) that correspond to the number of folds on the outer loop. The average of those scores results in an unbiased estimation of the performance of our algorithms. Note that a fold can contains both corrective or preventive maintenance. We target on an equal distribution of those two categories of runs across the different folds.

\section{Application to a rotating machine}
\label{sec:application}
\subsection{Problem description}
\begin{figure}[t]
    \centering
    \includegraphics[scale=0.3]{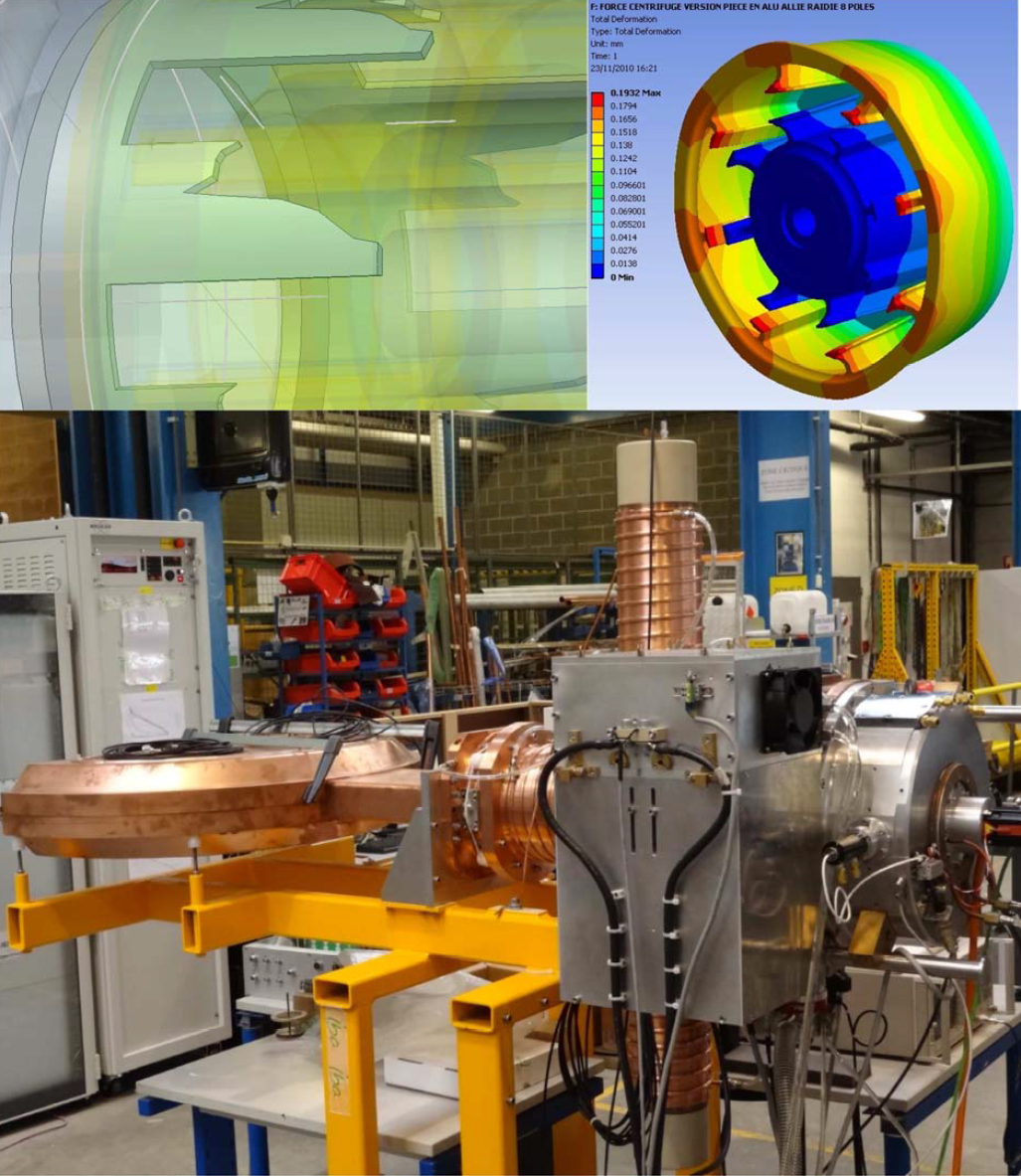}
    \caption{(bottom) RF system of the cyclotron with the rotating condenser on the right. (Top) Detailed view on the rotating condenser. \protect\cite{kleeven2013iba} CC-BY-3.0}
    \label{fig:s2c2_rf}
\end{figure}

\noindent The predictive maintenance case study we consider in this work deals with a high-speed rotating condenser (RotCo) modulating the RF frequency inside a synchrocyclotron \cite{kleeven2013iba}. The RotCo is composed of a stator and a rotor with eight blades and rotates at a constant speed of 7500 RPM with the help of ball bearings. A picture of the system is shown in Figure \ref{fig:s2c2_rf}.
Several sensors located inside the machine are used to gather data. An accelerometer sensor placed on the condenser external casing measures vibrations and performs 10-second acquisitions at a rate of 10kHz, once every hour. Four other sensors are placed on the machine to gather data every second. Those include two temperature sensors, a vacuum pressure and a torque sensor. In total, 11 corrective maintenance and 28 preventive maintenance runs have been gathered, i.e. a total of $R=39$ runs.\\
\\
After this data acquisition step, several features are built. For the vibration data, time-domain and frequency-domain features on each of the 10-second long acquisition files are constructed. Table \ref{tab:features} details the way those features have been computed.\\

\begin{table}[t]
    \centering
    \begin{tabular}{|m{0.25\textwidth}|m{0.35\textwidth}|m{0.25\textwidth}|}
    \hline
        \textbf{Raw signals} & \textbf{Feature extraction} & \textbf{Selected features} \\\hline
        \emph{Vibration amplitude} & \emph{Time-domain:}\newline RMS, MAD, Peak to Peak Amplitude, Skewness, Kurtosis, Crest Factor, Clearance Factor, Shape Factor, Margin Factor, Max Amplitude\newline \newline \emph{Frequency-domain:}\newline Amplitude 1N\footnotemark, 2N, 3N, every 20Hz band from 0-1kHz, BPFO 1-3N, BPFI 1-3N, BSF 1-3N, FTF 1-3N & MAD, Margin Factor, BPFO 3N, Peak to Peak Amplitude, Crest Factor, Amplitude 1N, RMS, Spectral Amplitude at 350 +- 10 Hz\\\hline
        \emph{Non-vibration features}: Bearing Temperature\newline Pyrometer Temperature \newline Torque \newline Vacuum Pressure & \emph{Time-domain:}\newline Mean, Max, Min, Standard Deviation, Skewness, Kurtosis & Torque Mean, Torque Max\\\hline
    \end{tabular}
    \caption{Feature design}
    \label{tab:features}
\end{table}

\noindent In the time-domain, those include Root Mean Square (RMS), Median absolute deviation (MAD) which is a robust measure of variability based on the deviations from the median, peak to peak values, skewness (third statistical moment) and kurtosis (fourth statistical moment). Those also include metrics based on the peaks of the signals: crest factor, clearance factor, shape factor, margin factor and max amplitude (for a detailed explanation on those features, the reader can refer to \cite{matlabfeatures}). For the frequency-domain features, we compute the amplitudes at the fundamental frequency and its first three harmonics, the spectral power of all 20 Hz non-overlapping bands from 0-1kHz and finally the amplitudes at characteristic bearing frequencies \cite{schoen1995motor}, i.e.
\footnotetext{1N refers to one time the fundamental frequency.}
\begin{itemize}
    \item Ball Pass Frequency Outer Race (BPFO): $\frac{n_f}{2}\left( 1 - \frac{D_b}{D_p}\cos\phi \right)$
    \item Ball Pass Frequency Inner Race (BPFI): $\frac{n_f}{2}\left( 1 + \frac{D_b}{D_p}\cos\phi \right)$
    \item Ball spin frequency (BSF): $\frac{D_pf}{2D_b}\left( 1 - (\frac{D_b}{D_p}\cos\phi)^2 \right)$
    \item Fundamental train frequency (FTF): $\frac{f}{2}\left( 1 - \frac{D_b}{D_p}\cos\phi \right)$
\end{itemize}
where $D_p$ is the pitch diameter, $D_b$ the ball diameter, $\phi$ the contact angle and $n$ the number of balls. The first three harmonics of those characteristic frequencies are also included. For the non-vibration data, aggregations of the signals over a 1-hour time-window are performed to match with the vibration acquisition sampling. Those aggregations include the mean, max, min, standard deviation, skewness and kurtosis values. This finally results in $N=89$ features (65 from vibration data and 24 from non-vibration data) computed every hour. The features are then scaled by subtracting their mean and dividing by their standard deviation, and then smoothed by a moving average over a 12-hour time window.\\
\\
The ten best features are then selected according to the feature selection method presented in \cite{hamaide2021unsupervised}. This is an unsupervised feature selection; therefore, no formulation is favored.\\
\\
All the formulations presented in Section \ref{sec:formulations} are tested on this real-world application. To avoid comparing all combinations of the approach of the first level with the ones of the second level, we first compare the approach of the first level with the simplest second level decision-making process, where the function $g$ is the identity function. This means that the health indicator is directly compared to a threshold and an alarm for replacement is raised whenever this output exceeds (or goes below) the optimized threshold. This allows us to select a subset of formulations obtaining the best scores, which are then tested against different aggregations $g$ in Section \ref{sec:lvl2}.\\
\\
The double cross-validation is performed as follows: 11 corrective maintenance runs are distributed among 11 folds and the 28 machines preventively replaced are distributed equally among those 11 folds. After leaving one fold aside as test set, we perform the inner loop of the cross-validation where the ML model is trained consecutively on all but one run (where a fold is defined as a run). The inner cross-validation allows to tune the hyperparameters of the model. Then the test set changes to the next outer fold and the whole process starts again.

\subsection{First level: training \& results}
\noindent Table \ref{tab:hyperparam} summarizes the different hyperparameters of the problem for each formulation as well as the different labelling scenarios tested. Note that the univariate model is absent from the table because no training is required at the first level of the framework.
\begin{table}[h]
    \centering
    \begin{tabular}{|m{0.25\textwidth}|m{0.35\textwidth}|m{0.25\textwidth}|}
    \hline
        \textbf{Formulation} & \textbf{Hyperparameters} & \textbf{Scoring metric} \\\hline
        Binary classification & $C=[10^{-2}, 10^{-1}, 1, 10]$ \newline $\gamma=[10^{-5}, 10^{-4}, 10^{-3}]$ \newline kernel: linear, RBF \newline horizon = [3,5,7,10] days & F-score\footnotemark \\\hline
        Multi-class classification & $C=[10^{-2}, 10^{-1}, 1, 10]$ \newline $\gamma=[10^{-5}, 10^{-4}, 10^{-3}]$ \newline kernel: linear, RBF \newline labelling 1: 3 classes (0-5 days, 5-10days, >10days) \newline labelling 2: 6 classes (0-2 days, 2-4days, ..., 8-10days, >10days) & Unweighted mean of F-score associated to each label \\\hline
        One-class SVM & $\nu=[0.01, 0.05, 0.1, 0.5]$ \newline $\gamma=[10^{-4}, 10^{-3}, 10^{-2}]$ \newline kernel: linear, RBF \newline Horizon = 15 days & F-score (with failure data included during testing)\\\hline
        Regression (RUL, RUL percentage, ReLu) & $C=[10^{-2}, 10^{-1}, 1, 10]$ \newline $\gamma=[10^{-5}, 10^{-4}, 10^{-3}]$ \newline $\epsilon=[0.01,0.1,0.5]$\newline kernel: linear, RBF \newline horizon for ReLu: $t_d=10$ days & Mean absolute error: $\text{MAE}(y, \hat{y}) = \frac{1}{n} \sum_{i=0}^{n_-1} \left| y_i - \hat{y}_i \right|.$ \\\hline
    \end{tabular}
    \caption{Training \& hyperparameter tuning for first level of the predictive maintenance framework}
    \label{tab:hyperparam}
\end{table}
\footnotetext{The F-score is defined as the harmonic mean between precision and recall, i.e. $F_{\text{score}} = \frac{2\times \text{precision}\times\text{recall}}{\text{precision}+\text{recall}}$}

\noindent For binary classification, several horizons are tested to split the data among healthy and unhealthy, ranging from 3 to 10 days. The F-score is used to select the best set of hyperparameters. For multi-class classification, two labelling scenarios are tested. The first one includes three classes defined in the following way: from 0 to 5 days prior to failure, 5 to 10 days, and more than 10 days. The second labelling scenario includes 6 classes defined as 2-day periods from the failure and a class defined as more than 10 days prior to failure. Hyperparameters are selected according to the mean of the F-score.\\
\\
For the One-Class SVM implementing anomaly detection, the model is only trained on data further than 15 days prior to failure for corrective maintenance and all data for preventive maintenance. To select the best hyperparameters set however, the model is tested on all data of a run and the hyperparameters that obtained the best F-score are selected. \\
\\
For the RUL and RUL percentage formulations, the training can only be done on corrective maintenance runs. For the ReLu formulation, preventive runs can be included, as the labels associated to those runs can be labelled as zero across their lifetime, since the ReLu is defined as nonzero only at a time $t_d$ prior to failure and monotically increasing until the actual failure. Parameter $t_d$ is fixed at 10 days in all our tests. The metric used to select the hyperparameters is the Mean absolute error for all regression formulations.\\
\\
The results of the double cross-validation for all formulations is outlined in Table \ref{tab:results_lvl1}. The count of true positives and false positives are summed across the different folds of the test set (every run is at least in one test set) and the $F_{\text{score}}$ of equation \eqref{eq:f_score} is computed with those counts with parameter $\beta=0.5$ to give more emphasis on avoiding false positives. The business score is averaged on all corrective maintenance runs since the business score is not defined for runs with preventive replacement. Finally, the final score is computed according to equation \eqref{eq:final_score} with parameter $\alpha=0.75$ to increase the emphasis on the $F_{\text{score}}$.\\

\begin{table}[]
\fontsize{9.5}{12}\selectfont
    \centering
    \begin{tabular}{|m{0.14\textwidth}|m{0.14\textwidth}|m{0.14\textwidth}|m{0.14\textwidth}|m{0.14\textwidth}|m{0.14\textwidth}|}
    \hline
        \textbf{Formulation} & \textbf{False Positives} & \textbf{True Positives} & \textbf{Business score} & \textbf{F score}& \textbf{Final score} \\\hline
        Univariate model & \textbf{2/39} & 8/11 & 0.436 & 0.894& 0.855 \\\hline
        One-class SVM & 16/39 & 5/11 & 0.257 & 0.556& 0.531 \\\hline
        Binary classification & 3 days: 6/39 \newline 5 days: 5/39 \newline 7 days: 4/39 \newline 10 days: \textbf{2/39} & 3 days: 9/11 \newline 5 days: 9/11 \newline 7 days: 9/11 \newline 10 days: 9/11 & 3 days: 0.495 \newline 5 days: 0.526 \newline 7 days: 0.465 \newline 10 days: 0.500& 3 days: 0.840 \newline 5 days: 0.860 \newline 7 days: 0.880 \newline 10 days: \textbf{0.919} &3 days: 0.811 \newline 5 days: 0.831 \newline 7 days: 0.845 \newline 10 days: \textbf{0.883}\\\hline
        Multi-class classification & 3 classes: 5/39 \newline 6 classes: 5/39 & 3 classes: 9/11 \newline 6 classes: 8/11 & 3 classes: 0.493 \newline 6 classes: 0.468 & 3 classes: 0.860 \newline 6 classes: 0.838 & 3 classes: 0.829 \newline 6 classes: 0.807 \\\hline
        RUL & 32/39 & 2/11 & 0.141 & 0.180 & 0.177 \\\hline
        RUL percentage & 12/39 & 6/11 & 0.254 & 0.657& 0.622 \\\hline
        ReLu & 6/39 & \textbf{10/11} & \textbf{0.617} & 0.858 & 0.837 \\\hline
    \end{tabular}   
    \caption{Results of first level. A run is considered a false positive if an alarm is raised more than 15 days in advance. An alarm is considered a true positive if it was raised between 0 and 15 days prior to failure in case of a run with failure. The business score is the mean of the business scores for all corrective maintenance. Bold values are the best results for each criterion. Final score is computed via equation \eqref{eq:final_score} with $\alpha=0.75$ and $\beta=0.5$ and is the score of interest that determines the best formulation.}
    \label{tab:results_lvl1}
\end{table}

\noindent Surprisingly, we observe that the simplest model, the univariate model based on the best feature with respect to the three prognostic metrics from \cite{hamaide2021unsupervised}, performs quite well, better than most of the approaches tested even though the other approaches also include this feature within their ten selected features. This feature is the median absolute deviation of the vibration amplitude averaged over the last 12 hours. It is the median of the absolute deviation from the data's median and is computed as $\text{MAD}(x)=\text{median}(|x_i - \bar{x}|)$ where $\bar{x}=\text{median}(x)$. \\
\\
In the second degree of refinement (see Figure \ref{fig:models}), the one-class SVM performs poorly with many false positives resulting in a low final score. For the third refinement, i.e. supervised learning, we can make several observations. In the binary classification, the more we increase the size of the faulty class, the better performance we obtain. This could be explained by two different reasons. The first one is that the class imbalance is reduced when the size of the faulty class increases. The second reason is that signs of faults already appear up to 10 days in advance. In the multi-class classification, performances are slightly lower than for the binary classification. Thus, splitting the runs into multiple classes does not seem to help.\\
\\
For the three regression formulations, the results are quite different from each other. Directly mapping the features input to the RUL does not seem to work at all. This could be explained by the fact that there is too much disparity between the life spans of the different runs, or simply not enough failure samples. This can also partly be explained by the fact that the $\epsilon$ parameter in the SVR formulation should be tuned more carefully for this type of formulation, since the labels are not scaled to 0-1 like for the RUL percentage or ReLu. The percentage of life formulation performs better than the conventional RUL but is still far behind the other formulations. This could also be explained by the fact that training is only performed on corrective maintenance for those two formulations. Finally, the ReLu formulation performs quite well although the number of false positive is high. \\
\\
In conclusion for this analysis of first-level formulations, we find that no method clearly outperforms all the others and that depending on which criteria we focus on (i.e. which column of Table \ref{tab:results_lvl1} we look at), several formulations can be recommended. The trade-off between high failure detection rate and low false alarm is one of those determining aspects. In our case study the binary classification formulation gives the best results in terms of the final score for a well-chosen window size.\\
\\
In the next section, different decision-making function $g$ are tested on the best algorithms obtained at the first level.

\subsection{Second level: training \& results}
\label{sec:lvl2}
\noindent In this section, we compare the application of a function $g$ on the computed health indicators of the first level that lead to the best scores, i.e. classification with a 10 days window for the faulty class, the multi classification with 3 classes, the ReLu formulation and the univariate model. We compare the case where $g$ is a moving average and an exponential smoothing. For all cases, we test a moving window of size 12-hours, 24-hours, 48-hours and 5-days. In the case of the exponential smoothing, since we have an infinite window, we use equation \eqref{eq:alpha} to match the parameter $\alpha$ with the window size. The results are outlined in Figure \ref{fig:results_lvl2} and a more detailed version is available in the appendix Tables \ref{tab:results_bc}-\ref{tab:results_un}. We observe that aggregating the computed HI with a moving average or exponential smoothing is not a guarantee for better results. However, by tuning the window size parameter, we are able to obtain better performance than the identity mapping for the exponential smoothing in all models. \\
\\
We observe that when the window size at the second level increases, the number of false alarms (false positives) decreases, but so as the number of cases detected (true positives) which sometimes translates into a lower score. The absolute minimization of false alarms might be wanted in some applications; thus, this second level of aggregation could be effective in that case. We also observe that the exponential smoothing performs better in general than the simple moving average. Therefore, using a scheme of decreasing weights with respect to time might be a good idea. The relatively modest impact of the function $g$ on the final results might also be due to the fact that the features inputs were also averaged across a 12-hour time-window before the model is applied.\\
\\
In the end, it is hard to give a definitive conclusion about the best formulation to use in a predictive maintenance scenario, due to the scarcity of runs and failure data, which greatly impacts the final score. However, some insights can still be taken. A simple univariate model can already be effective provided that a health indicator with high predictive power has been found. If we have corrective maintenance data, a supervised learning algorithm should lead to better performance than a one-class classification or anomaly detection algorithm. The binary classification formulation gives the best results when the time-window splitting healthy and faulty data is carefully selected. Finally, while directly mapping the RUL to the inputs is not a good idea, we find that a formulation such as the ReLu formulation which can be trained on both preventive replacements runs and runs-to-failure with a labelling mimicking an increasing fault severity starting at a certain time $t_d$ before the failure, results in good performance.

\begin{figure}
\centering
\begin{subfigure}{.5\textwidth}
  \centering
  \includegraphics[width=1\linewidth]{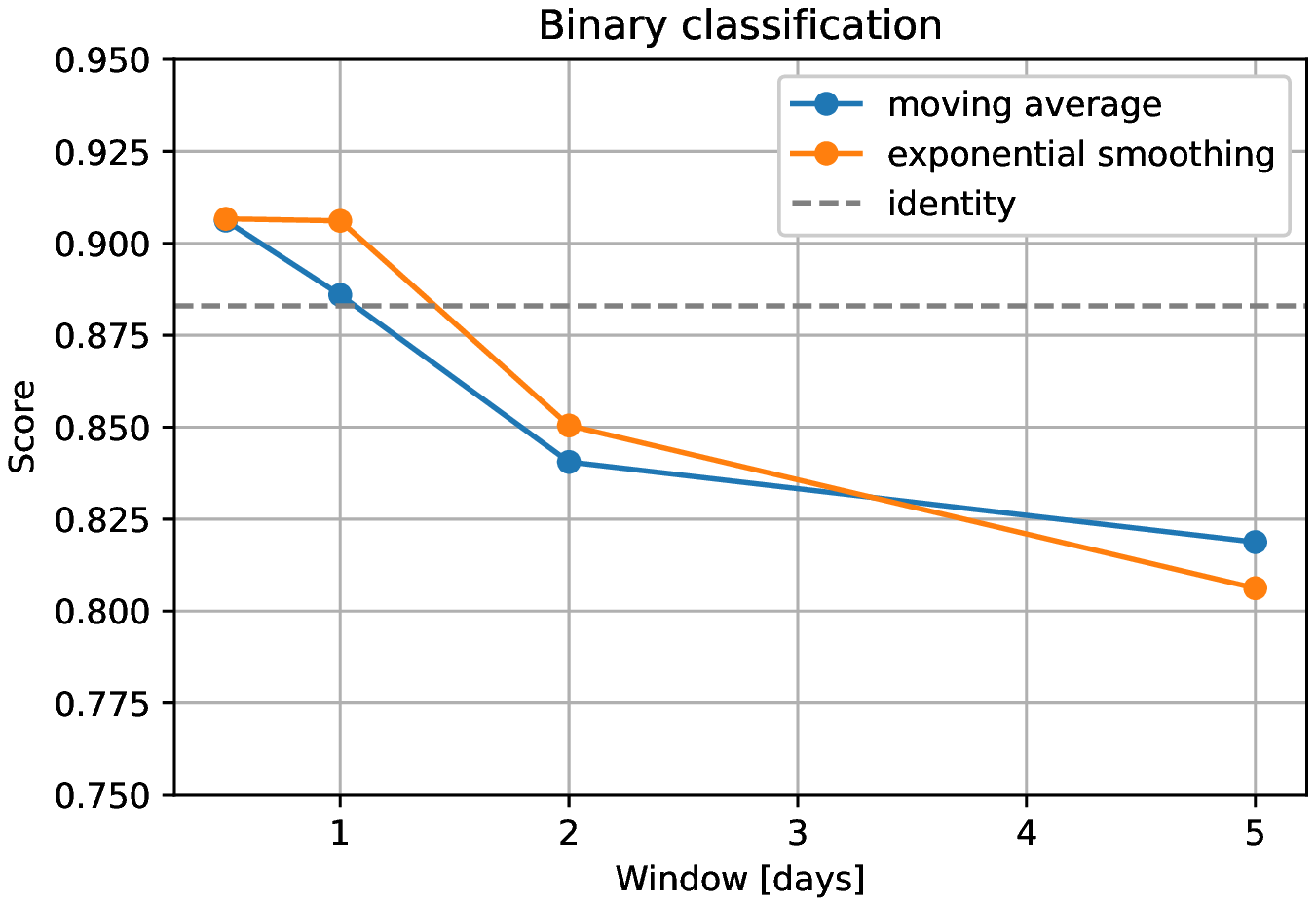}
\end{subfigure}%
\begin{subfigure}{.5\textwidth}
  \centering
  \includegraphics[width=1\linewidth]{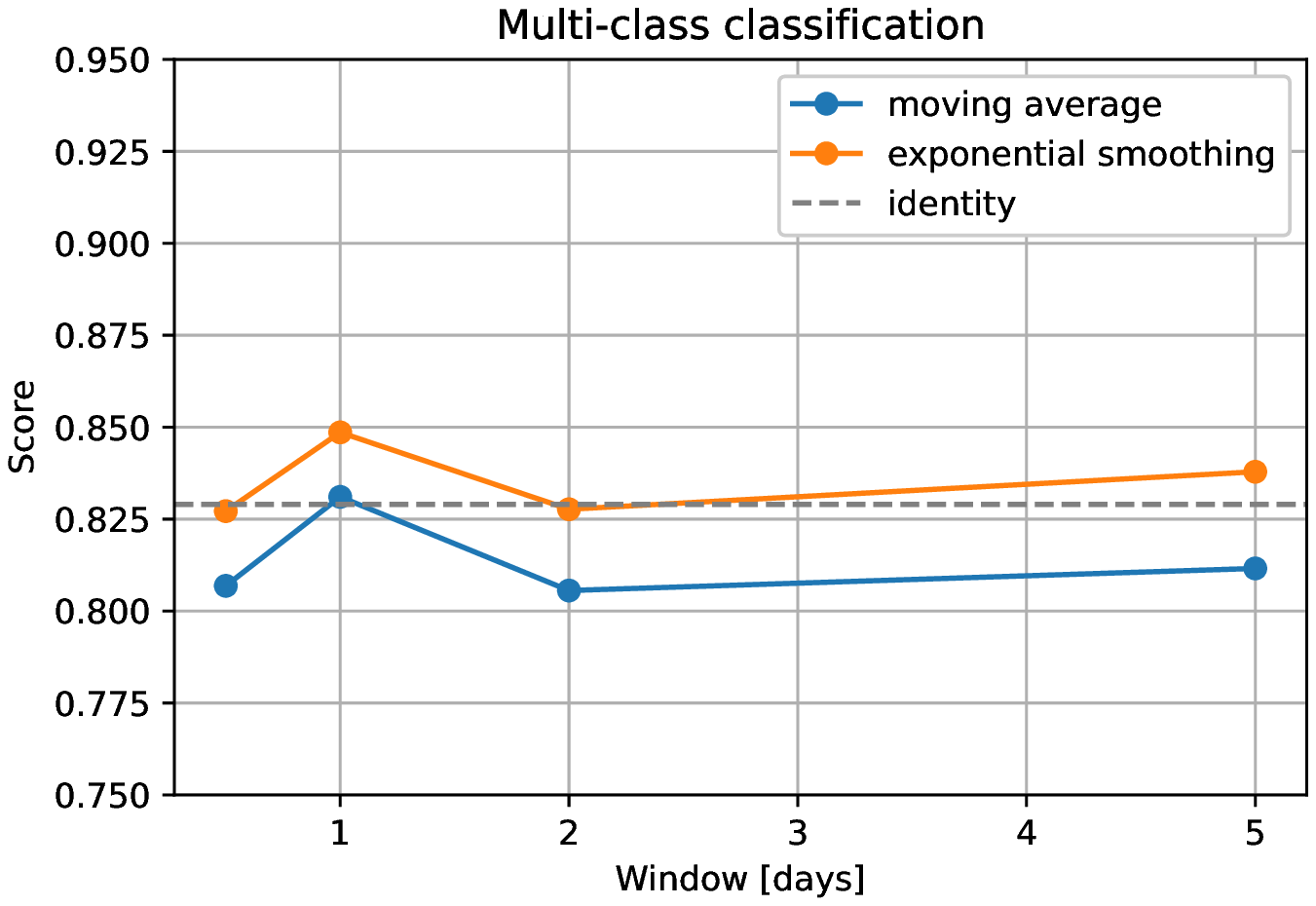}
\end{subfigure}
\begin{subfigure}{.5\textwidth}
  \centering
  \includegraphics[width=1\linewidth]{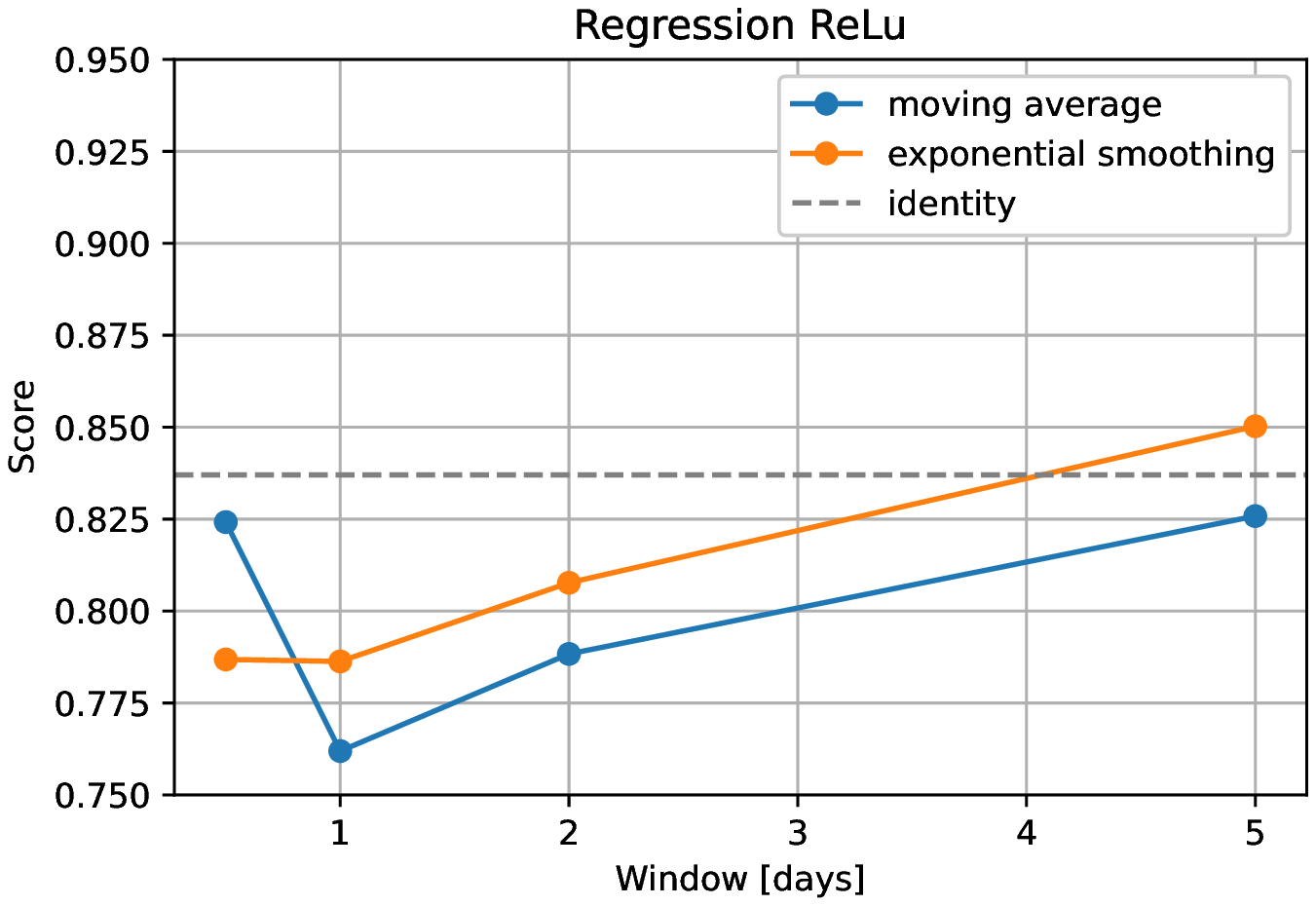}
\end{subfigure}%
\begin{subfigure}{.5\textwidth}
  \centering
  \includegraphics[width=1\linewidth]{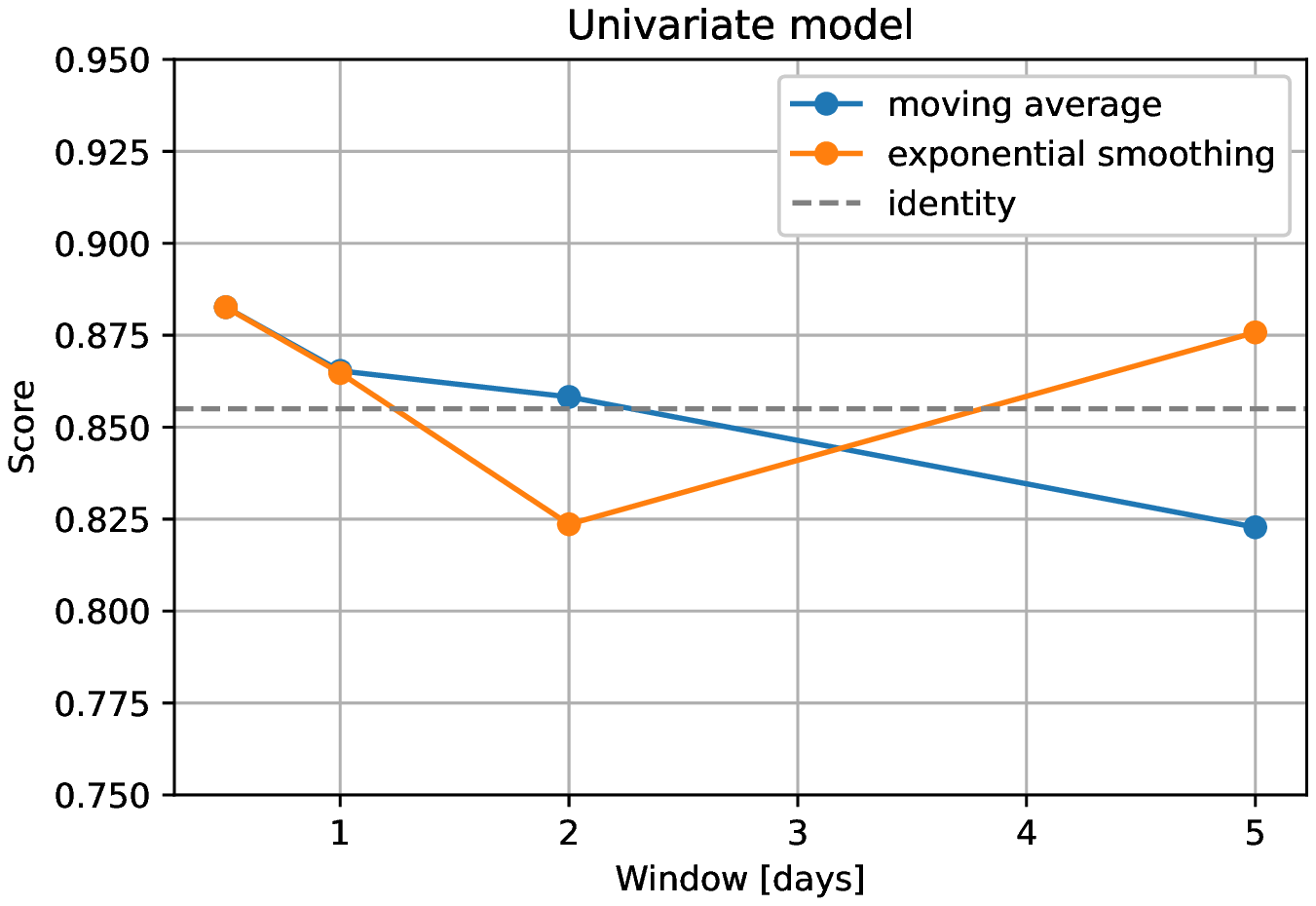}
\end{subfigure}
\caption{Results of second level}
\label{fig:results_lvl2}
\end{figure}

\section{Conclusion}
\label{sec:conclusion}
\noindent The aim of predictive maintenance is to avoid failure by replacing or repairing a machine at the right time. This is done by attempting to raise an alarm before the failure, and ideally sufficiently in advance to ease the maintenance scheduling. The best timing for detecting a failure was encoded via a business metric in our work. We developed a two-level framework to tackle this problem. In the first level, we compared several formulations to discriminate healthy and unhealthy states of the machines based on the remaining time before failure, in the absence of true labelled machine malfunctions. In the second level, we compared different ways to exploit the health indicator computed from the learning algorithm and we optimized a threshold for making the decision on when to raise an alarm. Our two-level framework approach gives promising results on the rotating machine case-study we considered. One of the key take-away is that the most complex models do not necessarily give the best results, and an univariate model can already perform very well when a powerful predictive feature can be first extracted. Although more complex multivariate models rely on more information and should theoretically lead to a more robust solution, it is not always the case. Special care must be taken to select the proper hyperparameters. Depending on the sought trade-off between false alarm and failure detection rate, certain methods and thresholds will perform better than other combinations, which should be carefully selected by the user and probably depend on the considered case study.

\bibliography{bibliography.bib}
\clearpage

\begin{table}[]
\begin{tabular}{|l|l|l|l|l|l|}
\hline
\textbf{Aggregation} & \textbf{FP} & \textbf{TP} & \textbf{B score} & \textbf{F score} & \textbf{Final score} \\\hline
MA 12h               & 2/39           & 10/11          & 0.54             & 0.940            & 0.906                \\\hline
EX 12h               & 2/39           & 10/11          & 0.546            & 0.940            & 0.906                \\\hline
MA 24                & 2/39           & 9/11           & 0.531            & 0.919            & 0.886                \\\hline
EX 24h               & 2/39          & 10/11          & 0.54             & 0.940            & 0.906                \\\hline
MA 48h               & 3/39          & 11/11          & 0.464            & 0.937            & 0.84                 \\\hline
EX 48h               & 4/39         & 9/11           & 0.532            & 0.880            & 0.85                 \\\hline
MA 5D                & 2/39           & 7/11           & 0.339            & 0.863            & 0.819                \\\hline
EX 5D                & 0/39           & 6/11           & 0.265            & 0.857            & 0.806       \\\hline
\end{tabular}
\caption{Binary classification results with moving average (MA) aggregation and exponential smoothing (EX)}
\label{tab:results_bc}
\end{table}
\begin{table}[]
\begin{tabular}{|l|l|l|l|l|l|}
\hline
\textbf{Aggregation} & \textbf{FP} & \textbf{TP} & \textbf{B score} & \textbf{F score} & \textbf{Final score} \\ \hline
MA 12h               & 5/39            & 9/11           & 0.438            & 0.860            & 0.824                \\ \hline
EX 12h               & 6/39            & 8/11           & 0.441            & 0.819            & 0.787                \\ \hline
MA 24                & 6/39            & 7/11           & 0.423            & 0.793            & 0.762                \\ \hline
EX 24h               & 6/39            & 8/11           & 0.435            & 0.819            & 0.786                \\ \hline
MA 48h               & 6/39            & 8/11           & 0.458            & 0.819            & 0.788                \\ \hline
EX 48h               & 5/39            & 8/11           & 0.48             & 0.838            & 0.808                \\ \hline
MA 5D                & 2/39            & 7/11           & 0.42             & 0.863            & 0.825                \\ \hline
EX 5D                & 0/39            & 7/11           & 0.348            & 0.897            & 0.85                 \\ \hline
\end{tabular}
\caption{Regression ReLu results}
\label{tab:results_relu}
\end{table}
\begin{table}[]
\begin{tabular}{|l|l|l|l|l|l|}
\hline
\textbf{Aggregation} & \textbf{FP} & \textbf{TP} & \textbf{B score} & \textbf{F score} & \textbf{Final score} \\ \hline
MA 12h               & 5/39            & 8/11           & 0.47             & 0.838            & 0.807                \\ \hline
EX 12h               & 5/39            & 9/11           & 0.472            & 0.860            & 0.827                \\ \hline
MA 24                & 5/39            & 9/11           & 0.518            & 0.860            & 0.831                \\ \hline
EX 24h               & 5/39            & 10/11          & 0.525            & 0.879            & 0.848                \\ \hline
MA 48h               & 5/39            & 8/11           & 0.455            & 0.838            & 0.805                \\ \hline
EX 48h               & 5/39            & 9/11           & 0.478            & 0.860            & 0.827                \\ \hline
MA 5D                & 5/39            & 8/11           & 0.525            & 0.838            & 0.811                \\ \hline
EX 5D                & 3/39            & 8/11           & 0.433            & 0.875            & 0.838                \\ \hline
\end{tabular}
\caption{Multi-class classification results}
\label{tab:results_mc}
\end{table}
\begin{table}[]
\begin{tabular}{|l|l|l|l|l|l|}
\hline
\textbf{Aggregation} & \textbf{FP} & \textbf{TP} & \textbf{B score} & \textbf{F score} & \textbf{Final score} \\ \hline
MA 12h               & 2/39            & 9/11           & 0.492            & 0.919            & 0.883                \\ \hline
EX 12h               & 2/39            & 9/11           & 0.492            & 0.919            & 0.883                \\ \hline
MA 24                & 3/39           & 9/11           & 0.496            & 0.9              & 0.865                \\ \hline
EX 24h               & 3/39            & 9/11           & 0.489            & 0.9              & 0.864                \\ \hline
MA 48h               & 2/39            & 8/11           & 0.475            & 0.894            & 0.858                \\ \hline
EX 48h               & 4/39            & 8/11           & 0.465            & 0.857            & 0.824                \\ \hline
MA 5D                & 2/39            & 7/11           & 0.385            & 0.863            & 0.822                \\ \hline
EX 5D                & 0/39            & 8/11           & 0.297            & 0.930            & 0.875                \\ \hline
\end{tabular}
\caption{Univariate model results}
\label{tab:results_un}
\end{table}

\end{document}